\def\ieeecopyrightnoteversion{}

\documentclass{article}
\usepackage{spconf,amsmath,graphicx}
\usepackage{xcolor}
\ifdefined\ieeecopyrightnoteversion
\usepackage{lastpage}
\usepackage{eso-pic}
\usepackage{calc}
\newlength{\msize}
\newlength{\mpwidth}
\fi

\title{Identity-guided Face Generation with Multi-modal Contour Conditions}

\name{Qingyan Bai$^{\dag}$ \qquad Weihao Xia  \qquad Fei Yin \qquad Yujiu Yang$^{\ddag}$}
\address{Tsinghua Shenzhen International Graduate School, Tsinghua University \\
bqy20@mails.tsinghua.edu.cn; 
\{xiawh3,feii.yin\}@outlook.com; yang.yujiu@sz.tsinghua.edu.cn}
\begin{document}
\maketitle
\def\thefootnote{\dag}\footnotetext{We sincerely appreciate Dr.Yujun Shen improving the manuscript.}\def\thefootnote{\arabic{footnote}}
\def\thefootnote{\ddag}\footnotetext{Acknowledgments. This work was supported by the Shenzhen Key Laboratory of Marine IntelliSense and Computation under Contract ZDSYS20200811142605016.}\def\thefootnote{\arabic{footnote}}
\begin{abstract}
Recent face generation methods have tried to synthesize faces based on the given contour condition, like a low-resolution image or sketch. However, the problem of identity ambiguity remains unsolved, which usually occurs when the contour is too vague to provide reliable identity information (\textit{e.g.}, when its resolution is extremely low). Thus feasible solutions of image restoration could be infinite. 
In this work, we propose a novel framework that takes the contour and an extra image specifying the identity as the inputs, where the contour can be of various modalities, including the low-resolution image, sketch, and semantic label map.
Concretely, we propose a novel dual-encoder architecture, in which an identity encoder extracts the identity-related feature, accompanied by a main encoder to obtain the rough contour information and further fuse all the information together.
The encoder output is iteratively fed into a pre-trained StyleGAN generator until getting a satisfying result.
To the best of our knowledge, this is the first work that achieves identity-guided face generation conditioned on multi-modal contour images. Moreover, our method can produce photo-realistic results with 1024$\times$1024 resolution.
\end{abstract}

\section{Introduction}
\label{sec:intro}

\begin{figure}[t]
\begin{center}
\includegraphics[width=0.75\linewidth]{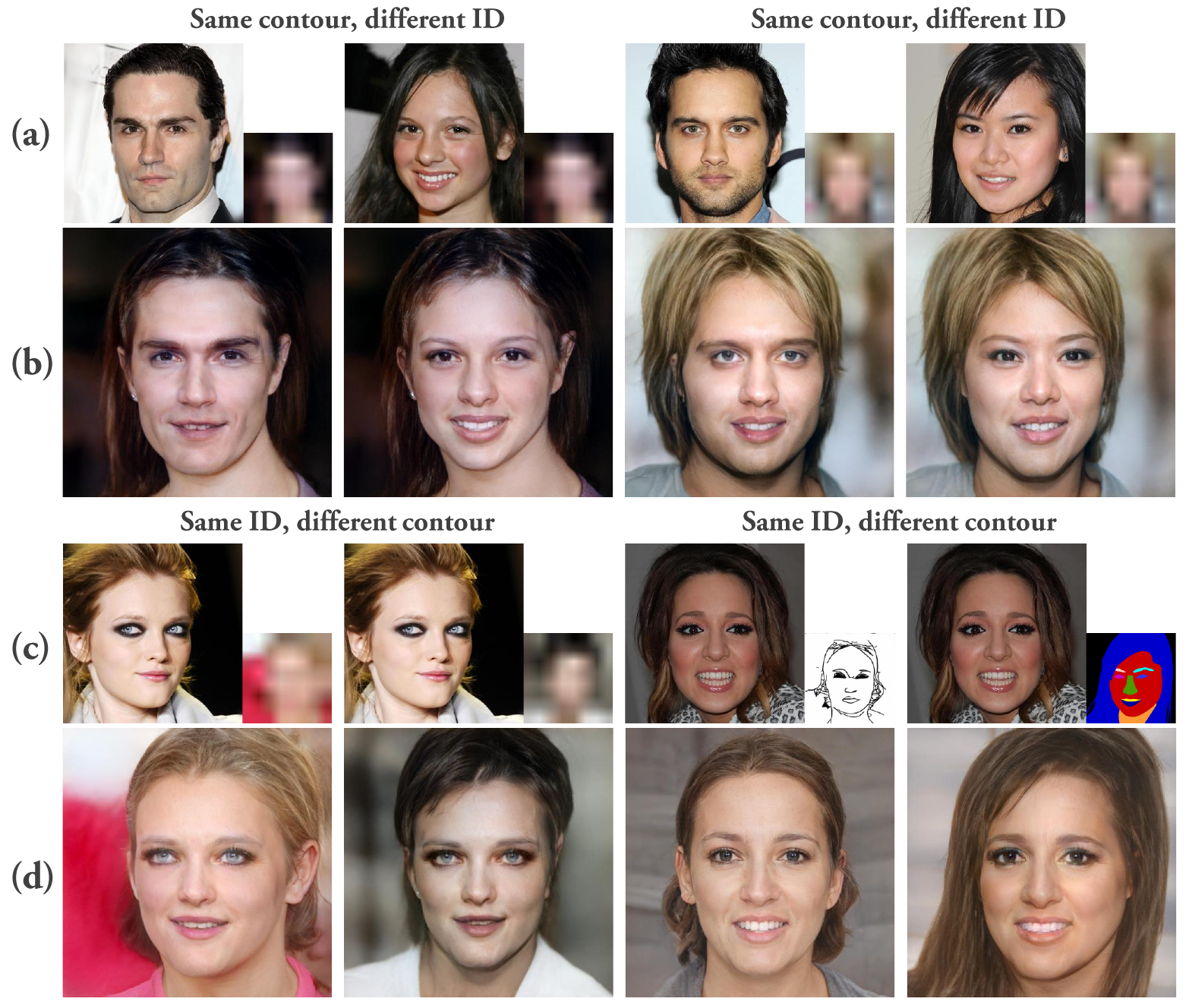}
\end{center}
\vspace{-5mm}
\caption{
Synthesized results ((b), (d)) of our model using the high-resolution images specifying \textbf{identities} and multi-modal \textbf{contour conditions} ((a), (c)).
}
\vspace{-5mm}
\label{fig:teaser}
\end{figure}

\begin{figure*}[t]
\begin{center}
\includegraphics[width=0.84\linewidth]{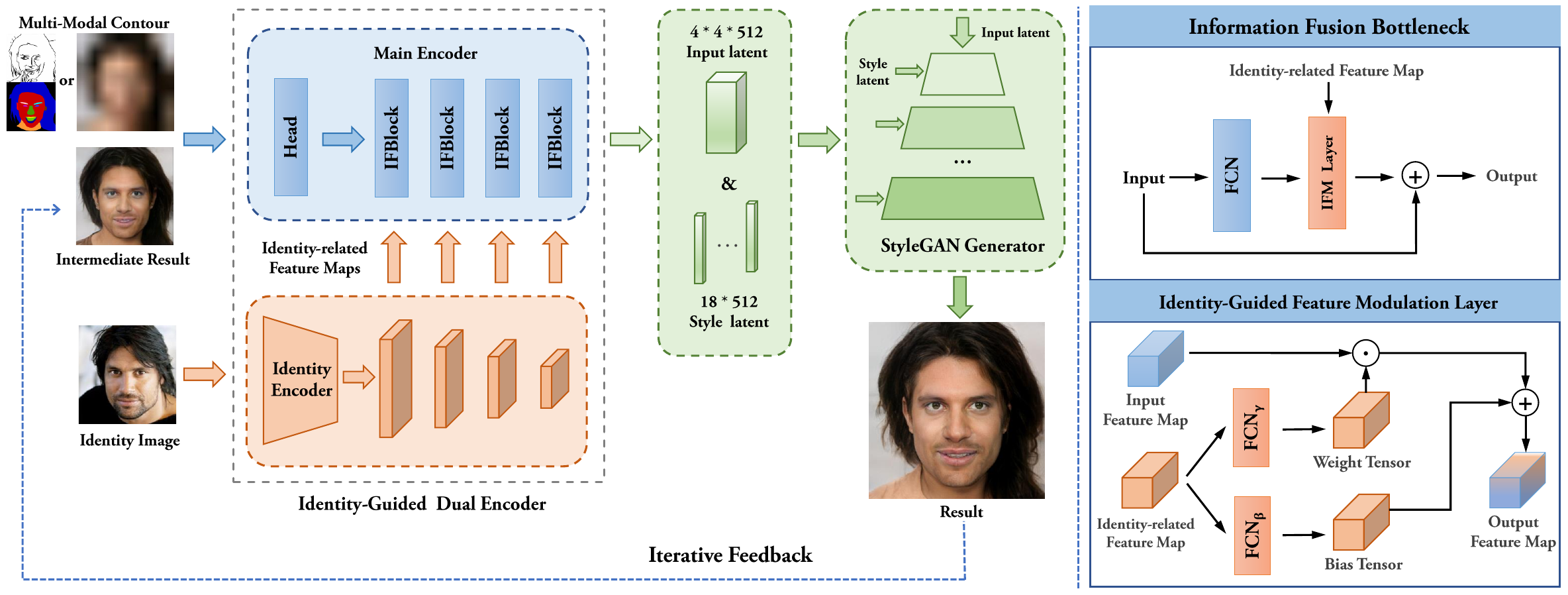}
\end{center}
\vspace{-5mm}
\caption{
The framework of our proposed method. The Identity-Guided Dual Encoder extracts identity information from an identity image and other information from a contour condition that can be of various modalities. 
The identity features are then injected into Information Fusion Blocks (IFBlocks), consisting of several Information Fusion Bottlenecks (IFBottlenecks), to fuse with contour information by Identity-Guided Feature Modulation (IFM) layers. The obtained input latent and style latent codes are then fed into corresponding layers of a pretrained StyleGAN to generate the desired results in an iterative manner.
}
\label{fig:pipeline}
\end{figure*}

In face synthesis and restoration tasks, identity (ID) ambiguity is a significant but unsolved problem. It usually happens when the input condition image is too vague to provide reliable identity information. 
For instance, given an audio~\cite{duarte2019speech,zhang2020apb2face,koumparoulis2020audio} or an extremely low-resolution (LR) face image~\cite{menon2020pulse,richardson2021pSp}, feasible solutions of image restoration could be infinite as the identity is too ambiguous to restore.  
This task especially fits the situation with requirements of synthesizing identity-specific rather than identity-agnostic faces, such as tracking known criminals.
Previous methods~\cite{menon2020pulse,richardson2021pSp} utilize a pretrained GAN to generate human faces based on low-resolution images. However, they did not fix the identity ambiguity problem and cannot generate faces with specified identities. 

In this paper, we propose a generic framework to generate a realistic face based on the identity face and contour, which could be of various modalities such as the low-resolution face, sketch, or mask (as in Fig.~\ref{fig:teaser}). We propose to use an identity encoder to obtain the multi-level identity-related feature and perform spatial-aware feature modulation to fuse the information and produce the desired latent codes.
To enlarge the capacity of latent space and preserve more spatial information, we encode the \textit{constant input} of StyleGAN synthesis network as the \textit{input latent code}. 
With the $\mathcal{L}_2$, LPIPS, identity, and $\mathcal{W}$-normalization losses, our model can be trained in a fully unsupervised manner and produces realistic and reasonable results with a resolution up to 1024$\times$1024.
A concurrent work~\cite{Mohamed2021lrguide} tries to migrate the attributes instead of the identity of the given high-resolution face. Due to the usage of the pretrained GAN rather than the traditional encoder-decoder framework, our model can produce realistic 1024$\times$1024 results while their resolution is limited to 256$\times$256.
Moreover, our method can utilize images of various modalities, such as the sketch and segmentation mask, while theirs can only handle low-resolution images.
The main contributions of this paper are as follows:

\begin{itemize}
\item We present a novel dual-encoder framework that is capable of generating 1024$\times$1024 realistic face images with desirable identity and low-resolution contour, which is a non-trival task.
    
\item 
To preserve the identity information, we propose to make spatial-aware feature modulation on multi-level identity-related features. 
Experiments reveal our model performs better than the state-of-the-art GAN inversion methods on the LR-conditioned task.
    
\item The proposed identity-guided method can also handle inputs of other modalities such as the sketch and mask, which proves the generalization ability of our approach.
\end{itemize}

\section{Proposed Method}
\label{sec:method}

\subsection{Preliminaries}
\label{subsec:pre}
StyleGAN~\cite{karras2019styleGAN} is a state-of-the-art unconditional image synthesis model, which includes a mapping network and a synthesis network.
The former maps a randomly sampled noise to a style latent code of 512 dimensions while the latter produces satisfying images with this latent code and a constant input by Adaptive Instance Normalization layers.
To deal with conditional synthesis tasks, recent methods~\cite{richardson2021pSp,abdal2019image2stylegan,zhu2020idinvert,gu2020multicode,bayat2021fast,yin2022styleheat} use a technique called GAN inversion~\cite{xia2021survey}. 
GAN inversion is to map an image into the latent space of a pretrained GAN model for a desired latent code, which can be faithfully reconstructed afterwards. 
We can then edit the given image by embedding additional information or discovering certain directions in the latent space.
Based on how the latent code is discovered, GAN inversion is categorized as optimization-based~\cite{abdal2019image2stylegan}, learning-based~\cite{richardson2021pSp}, or their combination~\cite{zhu2020idinvert}.
We refer readers to~\cite{xia2021survey} for a comprehensive survey.

\subsection{Overview}
\label{subsec:overview}
In the proposed task, the input includes a condition image $x_{c}$ and an identity image $x_{id}$. 
The input condition image (referred as contour conditions) can be of arbitrary modalities, such as the extremely low-resolution face, sketch, or semantic label, to provide the contour information.
Our goal is to synthesize a realistic face image whose identity and contour should be consistent with given identity image and contour condition.
Since we introduce a pretrained StyleGAN as the generator, the key problem becomes how to get an optimal latent code consistent with both the identity and the contour. 
To this end, we propose an Identity-Guided Dual Encoder for Face Generation, abbreviated as IDE.
As described in Fig.~\ref{fig:pipeline}, our IDE contains two encoders: the identity encoder and the main encoder. IDE produces a 18$\times$512 $\mathcal{W^{+}}$~\textit{style} latent code and an extra 4$\times$4$\times$512~\textit{input} latent code with spatial information. With the input latent code serving as the initialization and the style latent code serving as the modulation parameters of different layers, the pretrained StyleGAN generator can generate a satisfactory image after several iterations. 

\subsection{Identity-Guided Dual Encoder}
To deal with the identity ambiguity problem, we propose an Identity-Guided Dual Encoder, including an identity encoder and a main encoder.
The Resnet-based~\cite{he2016deep} identity encoder $E_{id}$ extracts the multi-level identity-related feature of the identity image. 
This process can be formulated as $\{F_{id}^{i}\}_{i=1}^{N} = E_{id}(x_{id})$, where $x_{id}$ and $F_{id}^{i}$ respectively represent the input identity image and the corresponding identity-related feature map extracted from the $i$-th block of the identity encoder. 
The main encoder takes the contour condition and the intermediate result as inputs, which is fed back from the last iteration by the StyleGAN generator. 
It contains a simple preprocessing convolutional head and $N$ (the same number as in the identity encoder) Information Fusion Blocks (IFBlocks).
Each IFBlock consists of several Information Fusion Bottlenecks (IFBottlenecks). 
IFBlocks extract the contour information from the contour condition and receive multi-level identity information from the corresponding blocks in the identity encoder. 
IFBottlenecks fuse the identity features with contour information of each level by Identity-Guided Feature Modulation (IFM) Layers.
In each IFM Layer, two Fully Convolutional Networks (FCNs)~\cite{long2015fully} are first adopted to convert the identity feature to modulation parameters, namely the weight tensor $\gamma$ and the bias tensor $\beta$:
\begin{equation}
\begin{aligned}
\gamma^{i,j} = {FCN}_{\gamma}^{j}(F^i_{id}), \\
\beta^{i,j} = {FCN}_{\beta}^{j}(F^i_{id}),
\end{aligned}
\end{equation}
where $i$ and $j$ denote the IFBlock index and the IFBottleneck index of the block, respectively.
Then the affine transformation is performed on the input feature map of $H^{i,j}$ with the weight and bias tensor to obtain the fused feature map activation of the next IFBottleneck $H^{i,j+1}$:
\begin{equation}
H^{i,j+1}_{c,y,x} = \gamma^{i,j}_{c,y,x} \times  H^{i,j}_{c,y,x} + \beta^{i,j}_{c,y,x}.
\label{eqn:IFM}
\end{equation}
To preserve the spatial information, $\gamma$ and $\beta$ are tensors instead of vectors as in~\cite{park2019SPADE} and $(c,y,x)$ indicates the site. 

To enlarge capacity of latent space and preserve spatial information, the last block of the main encoder predicts not only a classic 18$\times$512 $\mathcal{W^+}$ style latent code~\cite{abdal2019image2stylegan} but a 4$\times$4$\times$512 input latent code~\cite{zhu2021oneShotFaceSwapping}. By contrast with the classic constant input, which is a fixed tensor, the input latent code preserves more case-specific and spatial characteristics. 
Once the latents are obtained, the pretrained StyleGAN generator can output the desired face with them as described in Section~\ref{subsec:overview}.

\subsection{Loss Functions}
\label{subsec:loss}

This section describes the utilized loss functions in detail.
To preserve the low-frequency information of the contour image, we adopt pixel-wise $\mathcal{L}_2$ and LPIPS~\cite{zhang2018unreasonable} perceptual loss functions between the synthesized result $G(\text{IDE}(x_c, x_{id}))$ and the real face image $\hat{x_{c}}$ corresponding to the contour condition $x_{c}$, which can be formulated as:
\begin{equation}
\begin{aligned}
\mathcal{L_{\text{2}}} &= || \hat{x_{c}} - G(\text{IDE}(x_c, x_{id})) ||_2,\\
\mathcal{L_{\text{per}}} &= || P(\hat{x_{c}}) - P(G(\text{IDE}(x_c, x_{id}))) ||_2,
\end{aligned}
\end{equation}
where $\text{IDE}(\cdot)$ and $G(\cdot)$ respectively denote our proposed dual encoder and the pretrained StyleGAN generator, and $P(\cdot)$ denotes the perceptual feature extractor.

To preserve the original personal identity, we optimize the identity loss by calculating the cosine similarity between the identity feature of the output and identity images:
\begin{equation}
        \mathcal{L}_{id} = 1 - \frac{z_{id} \cdot {z}_{gen}}
        {\left \| z_{id} \right \| \left \| {z}_{gen} \right \|},
\end{equation}
where ${z}_{id}$ and ${z}_{gen}$ respectively denote the basis feature of the identity image and generated image obtained by ArcFace recognition model~\cite{deng2019arcface}.

We also find normalization of the latent codes plays a significant role in generating realistic faces. Specifically, we encourage the generating style latent vectors to be closer to the average latent vector to produce more reliable $\mathcal{W^+}$ latent codes. 
The $\mathcal{W}$ normalization loss is formulated as
\vspace{-1.15mm}
\begin{equation}
    \mathcal{L}_{\text{w}} = || w - \overline{w}||_2,
\end{equation}
where $\mathbf{w}$ denotes the output $\mathcal{W^+}$ style latent code of our encoder and $\overline{\mathbf{w}}$ denotes the average latent code obtained by random sampling. 
Our full objective function is defined as:
\vspace{-1.15mm}
\begin{equation}
    \mathcal{L}= 
    \lambda_1 \mathcal{L}_2 + 
    \lambda_2 \mathcal{L}_{\text{per}} + 
    \lambda_3 \mathcal{L}_{\text{id}} + 
    \lambda_4 \mathcal{L}_{\text{w}},
    \label{eqn:total_loss}
\end{equation}
where $\lambda_1$, $\lambda_2$, $\lambda_3$, and $\lambda_4$ are loss weights.

\section{Experiments}
\label{sec:exp}

\begin{figure}[t]
\begin{center}
\includegraphics[width=0.8\linewidth]{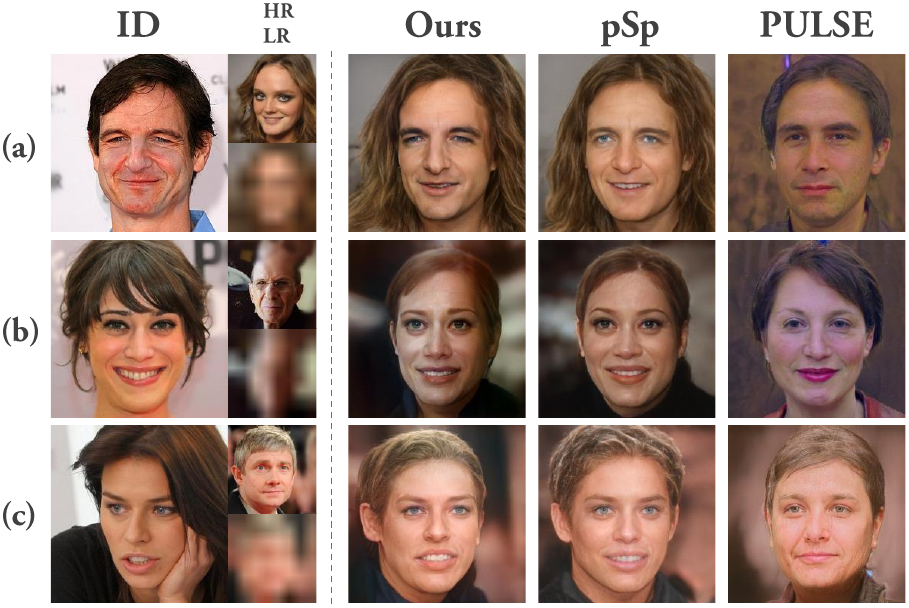}
\end{center}
\vspace{-5mm}
\caption{Comparison results with other methods subjected to the low-resolution contour. Three rows of comparison are illustrated to show that our method has advantages in (a) identity fidelity, (b) contour fidelity and (c) realism. Note that the original models of pSp and PULSE can not achieve this non-trivial task and we make changes to them.}
\vspace{-5mm}
\label{fig:comparison}
\end{figure}

\subsection{Implementation Details}
\noindent{\textbf{Datasets.}} We use CelebA-HQ~\cite{karras2018progressive} for training and evaluation, which contains 30,000 face images of 1024$\times$1024 resolution. 
For the multi-modal inputs, LR images are obtained by resizing the faces to 32$\times$32 with bicubic downsampling. Semantic labels and sketches are from~\cite{lee2020CelebAMaskHQ} and \cite{xia2021tedigan}, respectively.
For each contour condition (LR image, label or sketch), we randomly sample 10 identity images to obtain the contour-identity pairs. 
The original CelebA-HQ testing dataset contains 2,842 images. Thus the identity-guided testing dataset includes 28,420 contour-identity pairs. 

\noindent{\textbf{Training Setting.}}
Weighting factors in Equation~\eqref{eqn:total_loss} are set as $\lambda_1=0.1$, $ \lambda_2=1$,
$\lambda_3=0.5$, and $\lambda_4=0.003$.
Following the prior works~\cite{richardson2021pSp, alaluf2021restyle}, all the input image resolution is 256$\times$256, and the 1024$\times$1024 output image is resized to 256$\times$256 before calculating loss functions. 
We adopt the Adam optimizer~\cite{kingma2014adam} with a constant learning rate of $10^{-4}$. The batch size is set to 8. 
Since our method is the first high-fidelity identity-guided image translation task, we make some changes to pSp~\cite{richardson2021pSp} and PULSE~\cite{menon2020pulse} for fair comparisons.
For pSp, we concatenate the identity face and the contour as the input of the encoder and add the identity loss mentioned in Section~\ref{subsec:loss}.
For PULSE, we add the identity loss in the learning objective and rearrange the loss weights.

\begin{figure}[t]
\begin{center}
\includegraphics[width=0.75\linewidth]{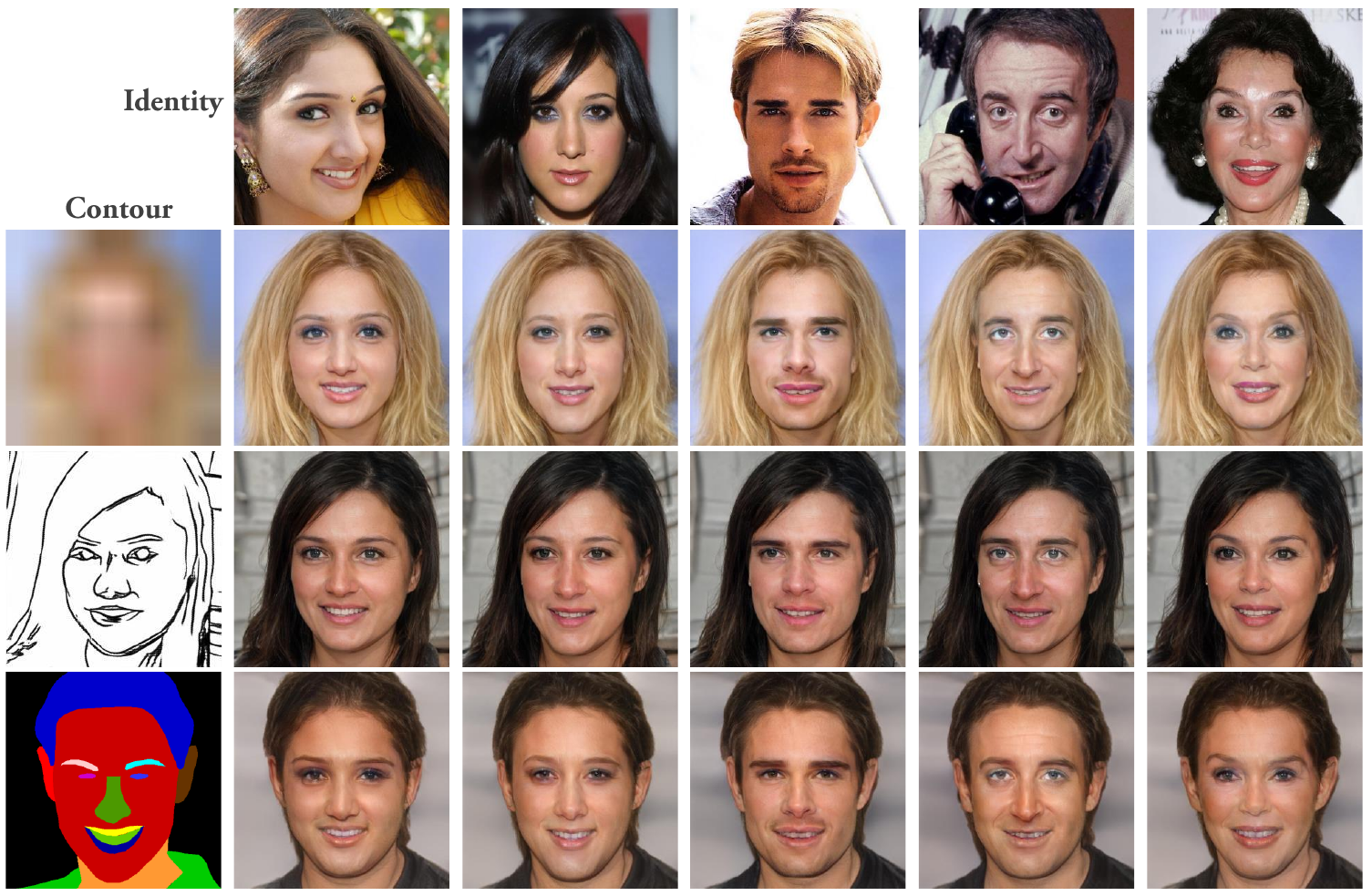}
\end{center}
\vspace{-5mm}
\caption{Results of multi-modal contour conditions.
}
\label{fig:multi-modal}
\vspace{-5mm}
\end{figure}

\subsection{Qualitative Results}

We qualitatively compare our method with two state-of-the-art GAN inversion based face synthesis methods pSp~\cite{menon2020pulse} and PULSE~\cite{richardson2021pSp}. 
As shown in Fig.~\ref{fig:comparison}, our method successfully generates faces with identity and contour preserved.
To some extent, both pSp and PULSE fail to preserve the original identity and contour of input images, especially in eyes and light effects. 
For realism, our model performs better in hair synthesis and PULSE often produces noise and artifacts even though it has lower FID.
Fig.~\ref{fig:multi-modal} shows more results for inputs of various modalities like the LR image, sketch or mask. As shown, our model can also produce results according to the contour condition while preserving the identity information.

\subsection{Quantitative Results}
As aforementioned, our method aims to combine the contour and the identity to synthesize a realistic high-resolution face.
We adopt LPIPS~\cite{zhang2018unreasonable} and FID~\cite{heusel2017gans} to evaluate image quality and diversity while using identity similarity (IDSIM) to evaluate identity similarity. 
LPIPS is calculated between the real high-resolution face image corresponding to the contour condition (\emph{e.g.}, the original HR image of the LR image) and the synthesized image to evaluate the fidelity. 
FID is adopted to evaluate realism and naturalism by calculating the distribution distance between the 28,240 testing output images and the 30,000 images from CelebA-HQ full dataset.
The identity similarity is calculated between the input identity face and the output face using a pretrained ArcFace~\cite{deng2019arcface} model. 

Table~\ref{tab:comparison} demonstrates the comparison between our method and three GAN inversion based face synthesis methods pSp~\cite{richardson2021pSp}, ReStyle~\cite{alaluf2021restyle} and PULSE~\cite{menon2020pulse} based on LR contours.
$\uparrow$ indicates the higher value of metric is better while $\downarrow$ means the opposite.
PULSE performs well in terms of LPIPS and FID but often fails to preserve the identity information probably because of its highly entangled $\mathcal{Z}$ latent space. 
Since it is optimization-based, it is also much more time-consuming compared with learning-based methods. 
Our method beats pSp in all three aspects thanks to the proposed dual-encoder architecture and the expanded latent space. 

\subsection{Ablation Study}
We present ablation studies based on the LR contours to reveal the necessity of each model component. 
Note that we choose ReStyle as our baseline rather than pSp as its encoder architecture is simpler and easier to modify. 
The `baseline' term in Table~\ref{tab:ablation} is our baseline that concatenates all three images as inputs and produces the $\mathcal{W^+}$ latent code with a single ReStyle encoder. 
`+IFBlock' indicates an extra identity encoder is added to the baseline to extract feature-related information and IFBlock is adopted to fuse information by applying spatial-aware feature modulation. We can conclude IFBlock effectively preserves the identity and improve the image quality. 
`+Input Latent' indicates that the input latent code, rather than the constant input, is fed to StyleGAN generator. The expanded latent space improves the synthesized image quality due to the enlarged spatial capacity.
The last row shows results of additionally loading parameters of a pretrained face recognition model to the identity encoder, which helps extract the identity-related feature more accurately.

\vspace{-5mm}
\begin{table}[th]
\small
\centering
\caption{Quantitative Comparison.}
\label{tab:comparison}
\begin{tabular}{cccc}
\hline
Method
& LPIPS$\downarrow$& IDSIM$\uparrow$& FID$\downarrow$\\
\hline
pSp~\cite{richardson2021pSp} &0.2911 &0.6067 & 67.9132\\
ReStyle~\cite{alaluf2021restyle} &0.2913 &0.4431 & 72.8556\\
PULSE~\cite{menon2020pulse}& 0.2758 & 0.0215 & \textbf{29.6793} \\
\textbf{Ours} & \textbf{0.2740}  & \textbf{0.7733} & 57.6614 \\
\hline
\end{tabular}
\end{table}

\vspace{-10mm}
\begin{table}[th]
\small
\caption{Ablation Study.}
\label{tab:ablation}
\centering
\begin{tabular}{cccc}
\hline
Method
& LPIPS$\downarrow$& IDSIM$\uparrow$& FID$\downarrow$\\
\hline
baseline& 0.2913& 0.4431& 72.8556\\
+IFBlock& 0.2862& 0.6681& 63.2523\\
+Input Latent& 0.2782& 0.7620& 60.8977\\
\textbf{Ours} & \textbf{0.2740}& \textbf{0.7733}& \textbf{57.6614}\\
\hline
\end{tabular}
\end{table}
\vspace{-5mm}

\section{Conclusion and Discussion}
\label{sec:conclusion}
In this work, we address the identity ambiguity problem in face synthesis and restoration.
We propose an identity-guided dual encoder that extracts identity features from an identity image and injects them into contour information from a multi-modal condition to acquire a desired latent code.
Our model can produce 1024$\times$1024 realistic faces that have the desirable contour and identity with no optimization required.

\noindent{\textbf{Ethical Consideration.}}
We strongly oppose the abuse of our method in violating privacy and security, considering its superior synthesis performance.  On the contrary, we hope it can be used to improve the existing fake detection systems.

\bibliographystyle{IEEEbib}
\bibliography{refs}

\end{document}